\documentclass[sigconf]{acmart}

\usepackage[T2A,T1]{fontenc}
\newcommand{\cyr}[1]{{\fontencoding{T2A}\selectfont #1}}

\setcopyright{none}
\renewcommand\footnotetextcopyrightpermission[1]{}

\begin{document}

\title[Post-OCR Correction of Mixed-Script Uzbek Text]{Post-OCR Correction of Mixed-Script Uzbek Text: Byte-Level ByT5 with an Empirical Homoglyph Noise Matrix}

\author{Mannon Ochilov}
\email{ochilov.mannon@mail.ru}
\affiliation{%
  \department{Department of Robotics and Intelligent Systems}
  \institution{Tashkent University of Information Technologies named after Muhammad al-Khwarizmi}
  \city{Tashkent}
  \country{Uzbekistan}}

\author{Oybek Narzullayev}
\email{o.o.narzullaev@gmail.com}
\affiliation{%
  \department{Department of Robotics and Intelligent Systems}
  \institution{Tashkent University of Information Technologies named after Muhammad al-Khwarizmi}
  \city{Tashkent}
  \country{Uzbekistan}}

\author{Bahodir Zaripov}
\email{amirbahodir@gmail.com}
\affiliation{%
  \department{Department of Artificial Intelligence}
  \institution{Tashkent State University of Economics}
  \city{Tashkent}
  \country{Uzbekistan}}

\renewcommand{\shortauthors}{Ochilov et al.}

\begin{abstract}
In Uzbek official and technical documents, a large share of OCR errors is caused not by poor image quality but by characters that are visually identical across the two alphabets in use. Because the script transition is still under way, Cyrillic and Latin appear on the same page and often within a single sentence, and in our measurements the density of homoglyph errors on such pages is close to five times higher than in single-script text. We study this problem in a setting where the corpus of engineering-geological and ICT infrastructure reports may not leave the organisation, so every stage was executed on local hardware. The proposed system combines a byte-level ByT5 model with a noise generator whose sampling weights are estimated from 359 manually classified error events, together with a Levenshtein-bounded conservative decoding rule. On a Gold set of 5,204 line pairs it lowers the character error rate from 0.0578 to 0.0369, restores 91.3\% of the homoglyphs and introduces only 84 new errors.
\end{abstract}



\maketitle
\pagestyle{plain}

\section{Introduction}
Optical character recognition has made it possible to search across large scanned collections~\cite{nguyen2021survey,smith2007}. Accuracy drops sharply, however, in low-resource and especially in multi-script settings, so whether a digital archive becomes usable at all is in many cases decided after recognition rather than during it~\cite{nguyen2021survey,chiron2017,rigaud2019}. Two alphabets, Cyrillic and Latin, are in parallel use, and in official and technical documents they mix on the same page and sometimes inside a single sentence. For an engine such as Tesseract~5.3~\cite{smith2007} this material is ambiguous on two levels at once: the glyphs are the same while the code points are not.

The course of the study was determined above all by the confidentiality of the documents. They are restricted engineering-geological reports and ICT records containing site coordinates, contract numbers and personal data. Three practical constraints follow. First, passing text fragments to commercial cloud APIs was not permitted, which excluded the baseline family most frequently cited in recent work~\cite{boros2024}; the usual reasons are recovery of memorised training data and the attack surface of hosted model services~\cite{carlini2021,yao2024}. Second, the Gold test set cannot be published, so reproducibility rests on aggregate statistics and an open noise generator~\cite{narzullaev2025repo}. Third, the entire pipeline had to run on a single machine.

Confidentiality also changed the weight attached to different kinds of error. Since no external auditor re-checks the output, the safety of the correction step matters more than its coverage. For this reason we treat the number of newly introduced errors, and not only the share of old errors removed, as the primary metric.

The characteristic defect of mixed-script text is the homoglyph error. Homoglyphs are characters that cannot be distinguished visually but carry different Unicode code points~\cite{gabrilovich2002,uts39,utr36}. Cyrillic \guillemotleft\cyr{а, е, о, р, с, х}\guillemotright{} and Latin \guillemotleft a, e, o, p, c, x\guillemotright{} are indistinguishable on the page, so the choice an operator makes between them is in practice arbitrary. A substitution such as \guillemotleft\cyr{ер}\guillemotright{}~$\rightarrow$~\guillemotleft ep\guillemotright{} or \guillemotleft\cyr{ВА}\guillemotright{}~$\rightarrow$~\guillemotleft BA\guillemotright{} breaks every downstream step that relies on string comparison, while the document still looks correct to a human reader.

The Gold set, never used for training or hyper-parameter selection and reserved for the final measurement alone, is the second element that shaped the work. To characterise the distribution empirically, its 359 homoglyph events were classified by hand (Table~\ref{tab:events}). On mixed-script pages the error density turned out to be almost five times that of single-script text. Most of the events occur at character level: 184 (51.3\%) single substitutions, 66 (18.4\%) clusters of adjacent characters and 77 (21.4\%) multi-position errors inside one token. In a further 32 cases (8.9\%) an entire word had migrated to the other alphabet, a pattern we refer to as polarisation.

\begin{table}[t]
  \caption{Classification of the 359 homoglyph events in the Gold test set}
  \label{tab:events}
  \begin{tabular}{lrr}
    \toprule
    Type of homoglyph event & Count & Share \\
    \midrule
    Single substitution & 184 & 51.3\% \\
    Cluster substitution & 66 & 18.4\% \\
    Whole-word polarisation & 32 & 8.9\% \\
    Multi-position substitution & 77 & 21.4\% \\
    \midrule
    Total & 359 & 100\% \\
    \bottomrule
  \end{tabular}
  \par\smallskip\footnotesize Shares are rounded, so the column may not sum to exactly 100\%.
\end{table}

An initial model trained on a general news corpus did not recognise the formatting conventions of specialised reports and corrupted words that were already correct (ERP $-111\%$). The paper therefore proposes a correction model built on the byte-level ByT5 architecture and an empirical homoglyph noise matrix, operating entirely inside the security perimeter.

\section{Related Work}
\paragraph{Dictionary- and statistics-based correction.}
The most common line of work treats the task as dictionary-based spell checking: candidates are collected by edit distance~\cite{levenshtein1966} and then re-ranked with a lexicon, a confusion matrix and an $n$-gram model. Nguyen et al.~\cite{nguyen2021survey} survey this direction, and their study of error statistics~\cite{nguyen2019deep} shows that errors depend heavily on the collection, so a confusion matrix estimated on one corpus transfers poorly to another. The ICDAR competitions~\cite{chiron2017,rigaud2019} gave the field a common evaluation protocol, and Amrhein and Clematide~\cite{amrhein2018} demonstrated that character-level statistical machine translation is still a strong baseline. A lexicon-driven system assumes that a corrupted word is a word missing from the dictionary, and homoglyph errors break exactly that assumption.

\paragraph{Neural post-correction and byte-level representation.}
Neural post-correction commonly relies on byte- or character-level representations and treats OCR errors as monolingual translation from noisy into clean text. Earlier studies established the effectiveness of character-level NMT~\cite{hamalainen2019}, showed that separating detection from correction reduces over-correction~\cite{schaefer2020}, and reported that BERT-based models outperform statistical pipelines~\cite{soper2021,lewis2020}. Others noted the drop in performance when the domain changes~\cite{todorov2020}, the advantage of lexically aware methods under low-resource conditions~\cite{rijhwani2021}, and cases in which an ensemble of small models beats a single large one~\cite{ramirezorta2022}. Two conclusions carry over to our setting: when the domain shifts, the dominant failure mode is over-correction~\cite{schaefer2020,todorov2020}, and for orthographic noise a character-level input is the most effective choice~\cite{hamalainen2019,ramirezorta2022}.

The difficulty begins at representation. Most transformer models~\cite{vaswani2017} use sub-word tokenisation, which is fragile under character-level errors: one changed character can make the tokeniser split a word into a completely different sequence. CANINE~\cite{clark2022canine} works directly on Unicode code points and Charformer~\cite{tay2022charformer} learns its own segmentation, while ByT5~\cite{xue2022byt5} adapts T5~\cite{raffel2020} to operate on raw UTF-8 bytes and is therefore markedly more robust to noise.

\paragraph{Homoglyphs, Unicode security and Uzbek resources.}
Homograph attacks were described early on~\cite{gabrilovich2002}, and the Unicode Consortium later standardised the detection of confusable characters in UTS~\#39 and UTR~\#36~\cite{uts39,utr36}. UTS~\#39 treats a mixed-script string as a risk signal, whereas in bi-alphabetic Uzbek text such strings are perfectly ordinary, so conventional rules do not discriminate well enough in practice.

A number of resources have been created for Uzbek, yet most of them are designed for a single alphabet. UzBERT~\cite{mansurov2021} was among the first monolingual models; cross-lingual embeddings for Turkic languages~\cite{kuriyozov2020} and a semantic similarity benchmark~\cite{salaev2022} have also been produced. Multilingual models such as XLM-R~\cite{conneau2020} and mT5~\cite{xue2021mt5} cover Uzbek but allocate it limited resources, particularly for Cyrillic. Work on Uzbek speech recognition~\cite{mukhamadiyev2023} and on punctuation restoration~\cite{musaev2025punct} points to the same need for in-domain adaptation.

The closest study to ours corrects Uzbek spelling errors with machine learning~\cite{ochilov2025spell}, but its hybrid algorithm relies on lexical errors, and homoglyph substitutions pass a dictionary check unnoticed. Uzbek question-answering~\cite{khujayarov2025qa} and sign-language translation~\cite{musaev2024sign} likewise operate on tokenised text and cannot detect an alphabet switch inside a token. Existing Uzbek corpora are almost always Latin-only or Cyrillic-only; mixed-script corpora are practically absent.

\paragraph{Large language models and the confidentiality barrier.}
Boros et al.~\cite{boros2024} tested LLMs on historical transcripts and observed improved fluency alongside a loss of fidelity to the source, with paraphrase and hallucination. Our own experiments with Qwen2.5-7B~\cite{qwen2024} gave an even weaker result on technical strings such as \guillemotleft\cyr{КМК} 2.01.03-19\guillemotright{} or \guillemotleft LMS.TUIT.UZ\guillemotright, because units of this kind carry no linguistic redundancy and the model is strongly inclined to \textquotedblleft{}correct\textquotedblright{} them incorrectly.

\section{Methodology}
\subsection{Corpus, confidentiality and ethics}
The raw corpus consists of 487,205 lines. After cleaning by length and by corrupted characters, the operational corpus came to 422,815 lines. On this basis 2,283,198 synthetic training pairs, 63,420 validation pairs and 63,426 test pairs were derived, which corresponds to about five noisy variants per source line. Since synthetic data reflect only the noise model that produced them, a separate Gold set was collected from genuine scanned documents. A line-by-line review discarded 849 of the initial 6,053 pairs, leaving a final Gold set of 5,204 pairs. Its distribution over alphabets can be written as follows:
\begin{equation}
  D_{\mathrm{gold}} = D_{\mathrm{cyr}} \cup D_{\mathrm{lat}} \cup D_{\mathrm{mix}}
\end{equation}

\subsection{Empirical noise generation}
The proposed generator does not assume a noise distribution; it estimates one directly and empirically from the scanned documents, which is what distinguishes it from the usual practice of injecting uniform noise at character level. The proportions given in Table~\ref{tab:events} (51.3\%, 18.4\%, 8.9\% and 21.4\%) are used as the sampling probabilities of the generator. Homoglyph candidates are taken from the Unicode confusables tables~\cite{uts39} and then re-weighted according to the frequencies actually observed, and each clean character is mapped onto its corresponding set of homoglyphs. The substitution is applied in line with these probabilities:
\begin{equation}
  \tilde{c}_i =
  \begin{cases}
    h_j \in H(c_i), & \text{with probability } p,\\
    c_i, & \text{with probability } 1-p,
  \end{cases}
\end{equation}
where $H(c_i)$ is the confusable set of the character $c_i$ and $p$ is the substitution rate. The value of $p$ is not fixed beforehand but calibrated so that the generated error density matches the density measured on the scans.

Whole-word polarisation, observed 32 times in the Gold set, was unexpected but important. A character-level model cannot reproduce it, because choosing each character independently yields mixed-script tokens, whereas in practice the tokens are entirely Latin or entirely Cyrillic. A token-level parameter was therefore added: it selects an alphabet for the whole token and applies that choice to all of its characters:
\begin{equation}
  \tilde{t} =
  \begin{cases}
    \pi(t), & \text{with probability } q,\\
    (\tilde{c}_1 \ldots \tilde{c}_n), & \text{with probability } 1-q.
  \end{cases}
\end{equation}
Here $\pi(\cdot)$ maps every character of the token to the opposite alphabet and $q$ denotes the probability of whole-word polarisation. The clearest example in the data is the token \guillemotleft PoE\guillemotright, which occurs in the scans in fully Latin, fully Cyrillic (\guillemotleft\cyr{РоЕ}\guillemotright) and mixed form.

\subsection{Architecture, conservative decoding and comparison systems}
The task was formulated as a byte-level sequence-to-sequence mapping and optimised on top of ByT5~\cite{xue2022byt5}. Because the model works directly with UTF-8, it represents homoglyph substitutions as what they actually are, namely edits at the byte level.

Since domain mismatch makes a model prone to over-correction, a conservative mechanism was introduced at the decoding stage. If the Levenshtein distance~\cite{levenshtein1966} between the prediction and the input exceeds a fixed threshold, the system rejects the model output and keeps the original line:
\begin{equation}\label{eq:conservative}
  y_{\mathrm{final}} =
  \begin{cases}
    \hat{y}, & \text{if } d(\tilde{x}, \hat{y}) \le \tau,\\
    \tilde{x}, & \text{otherwise.}
  \end{cases}
\end{equation}
The threshold was selected on the validation set as $\tau = 3$ characters, which permits one or two homoglyphs to be repaired but prevents large-scale rewriting.

mT5~\cite{xue2021mt5}, mBART-50~\cite{tang2020mbart} and CANINE~\cite{clark2022canine} were also examined but not included in the comparison experiments. This is not a judgement about their accuracy; the research question was how much a byte-level representation gains from an empirical noise distribution, which needs no broad comparison of multilingual architectures, and byte-level models are already known to be more stable under character-level corruption~\cite{xue2022byt5,clark2022canine,tay2022charformer}.

Two practical constraints also applied: a candidate model had to run on local resources, and the whole pipeline had to fit within a single GPU. ByT5 was therefore chosen as the neural component, with a rule-based corrector and a local LLM as the deployable alternatives.

\subsection{Metrics and success criteria}
The first metric is the character error rate, where $S$, $D$ and $I$ are the numbers of substituted, deleted and inserted characters and $N$ is the number of characters in the reference text:
\begin{equation}
  \mathrm{CER} = \frac{S + D + I}{N}
\end{equation}
The second is the error reduction percentage, which we report as the main result because it shows how much of the existing error the system actually removes:
\begin{equation}
  \mathrm{ERP} = \frac{\mathrm{CER}_{\mathrm{raw}} - \mathrm{CER}_{\mathrm{model}}}{\mathrm{CER}_{\mathrm{raw}}} \times 100\%
\end{equation}
Table~\ref{tab:systems} additionally reports two further indicators. The homoglyph restoration rate is the share of the 359 annotated events for which the correct code point was recovered. Newly introduced errors are counted at character level: a character that matched the reference in the input but was changed in the model output counts as a new error.

The acceptance criteria were agreed with the document owner at the beginning of the study and were not revised afterwards. Criterion M1 requires the number of errors on the Gold set to be reduced by at least 15\%. Criterion M2 requires that no more than 100 new errors be introduced across the whole set.

\section{Results}
\subsection{Training dynamics}
\begin{figure*}[t]
  \centering
  \includegraphics[width=\textwidth]{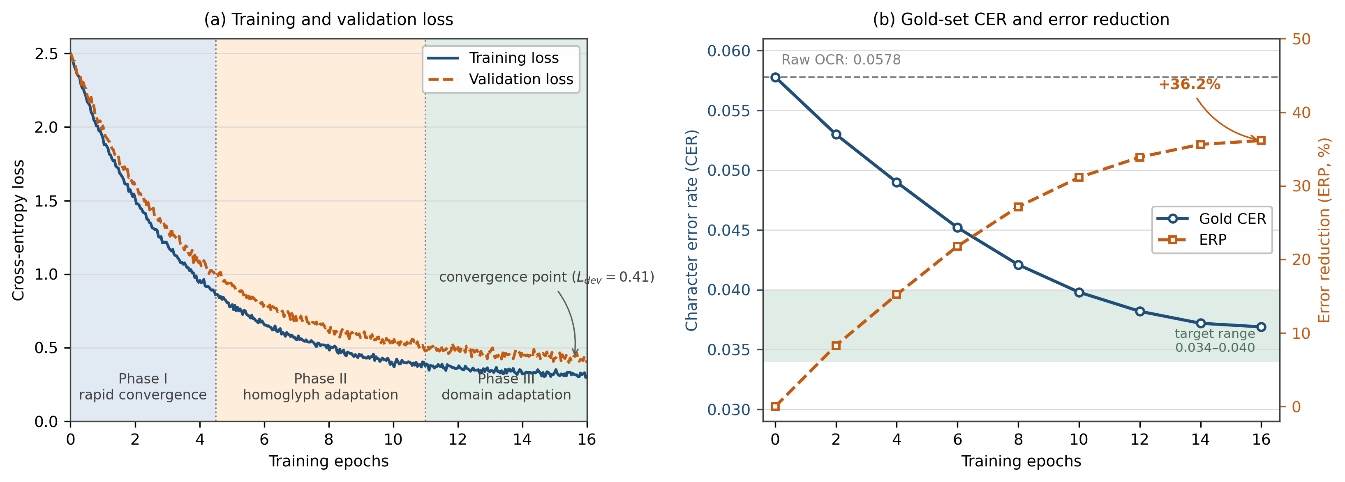}
  \caption{Training dynamics over 16 epochs: (a) training and validation cross-entropy loss; (b) CER and error reduction on the Gold set, evaluated every second epoch.}
  \Description{Left panel: training and validation cross-entropy loss falling from about 2.5 to about 0.3 to 0.4 over 16 epochs, with three shaded training phases. Right panel: Gold-set CER decreasing from 0.0578 to 0.0369 while error reduction rises to 36.2 percent.}
  \label{fig:training}
\end{figure*}

Training proceeded in three phases, with the composition of the data changed at epochs 4, 5 and 11. The shaded regions in Figure~\ref{fig:training}(a) mark these boundaries. Validation loss stayed slightly above training loss but showed no upward trend, indicating no overfitting. Figure~\ref{fig:training}(b) shows the Gold set: CER starts from the raw OCR value of 0.0578, falls steeply over the first six epochs, then stabilises at 0.0369, while ERP rose to +36.2\%.

\subsection{System comparison}
All systems were evaluated on the same Gold set of 5,204 lines (Table~\ref{tab:systems}).

\begin{table}[t]
  \caption{Comparison of the systems on the Gold test set (5,204 pairs)}
  \label{tab:systems}
  \small
  \setlength{\tabcolsep}{4pt}
  \begin{tabular}{lrrrr}
    \toprule
    System & CER$\downarrow$ & ERP$\uparrow$ & Homogl.$\uparrow$ & New err.$\downarrow$ \\
    \midrule
    Raw OCR & 0.0578 & -- & 0 & 0 \\
    Rule-based baseline & 0.0575 & +0.5\% & 7.2\% & 38 \\
    LLM (Qwen2.5-7B) & 0.0534 & +7.6\% & 52.4\% & 1,132 \\
    ByT5, no domain adapt. & 0.0491 & +15.1\% & 69.8\% & 286 \\
    \textbf{ByT5, proposed} & \textbf{0.0369} & \textbf{+36.2\%} & \textbf{91.3\%} & \textbf{84} \\
    \bottomrule
  \end{tabular}
  \par\smallskip\footnotesize Raw OCR is Tesseract 5.3~\cite{smith2007}. The LLM was run from a local checkpoint. The proposed system satisfies both criteria: ERP +36.2\% (M1 threshold +15\%) and 84 new errors (M2 threshold 100).
\end{table}

Only the last row of Table~\ref{tab:systems}, the proposed ByT5-based system, improves on both criteria at the same time. Compared with Qwen2.5-7B~\cite{qwen2024}, the proposed system removed 4.8 times more errors and introduced 13.5 times fewer new ones.

The rule-based corrector shows how much the noise model matters: the same algorithm restored 99.3\% of the homoglyphs on the synthetic test but only 7.2\% on the Gold set, ERP falling from +20.1\% to +0.5\%, with the corrector unchanged and only the noise distribution differing. The ByT5 variant without domain adaptation almost satisfies M1 on its own yet fails M2 with 286 new errors; both criteria were met together only after domain adaptation and conservative decoding were added.

\begin{figure*}[t]
  \centering
  \includegraphics[width=\textwidth]{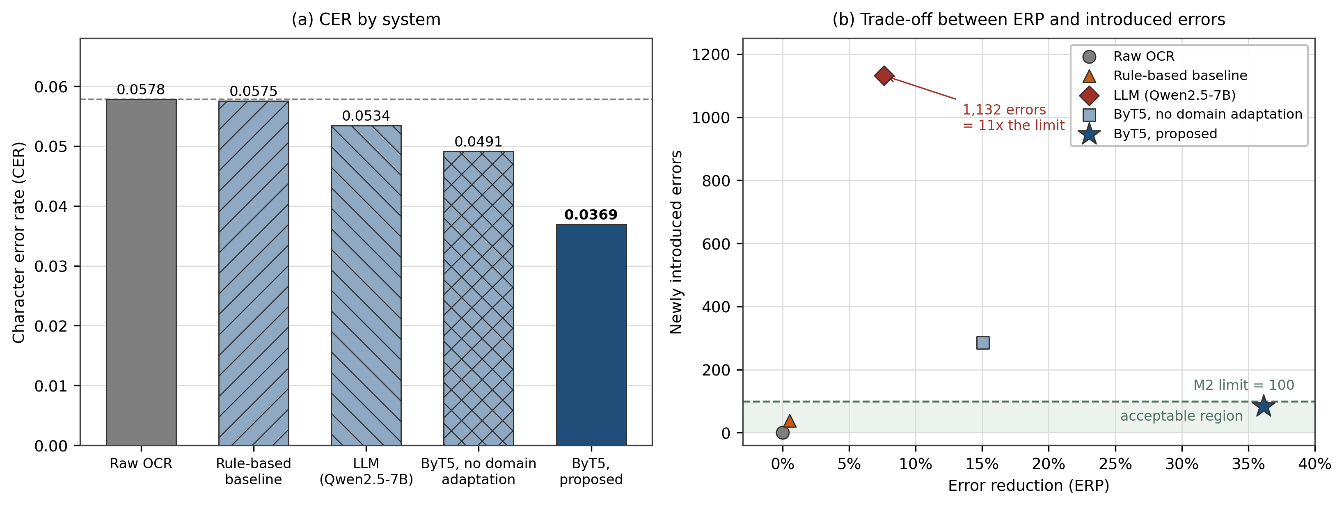}
  \caption{System comparison: (a) CER across systems; (b) the trade-off between error reduction and newly introduced errors, with the M2 threshold set at 100.}
  \Description{Left panel: bar chart of character error rate for raw OCR (0.0578), rule-based baseline (0.0575), Qwen2.5-7B (0.0534), ByT5 without domain adaptation (0.0491) and the proposed ByT5 (0.0369). Right panel: scatter of error reduction against newly introduced errors; only the rule-based and proposed systems lie below the limit of 100 new errors.}
  \label{fig:comparison}
\end{figure*}

Figure~\ref{fig:comparison}(b) makes this balance visible: the rule-based and the proposed systems lie inside the acceptable region, the variant without domain adaptation sits slightly above it, and the LLM falls far outside on the introduced-error axis for very little additional ERP.

\subsection{Qualitative analysis of corrections}
The aggregate metrics state how much error the system removes but not how it behaves. Table~\ref{tab:examples} places the two characteristic cases side by side: homoglyph substitutions that are repaired, and technical strings that are deliberately left alone. The examples are drawn from strings already cited in this paper, so no new material from the Gold set is disclosed.

\begin{table*}[t]
  \caption{Representative examples of model behaviour}
  \label{tab:examples}
  \small
  \setlength{\tabcolsep}{4pt}
  \begin{tabular}{lllll}
    \toprule
    Raw OCR & Model output & Reference & Code points (raw $\rightarrow$ model) & Model behaviour \\
    \midrule
    ep (Lat.) & \cyr{ер} (Cyr.) & \cyr{ер} (Cyr.) & \texttt{U+0065 U+0070} $\rightarrow$ \texttt{U+0435 U+0440} & Homoglyph restored \\
    BA (Lat.) & \cyr{ВА} (Cyr.) & \cyr{ВА} (Cyr.) & \texttt{U+0042 U+0041} $\rightarrow$ \texttt{U+0412 U+0410} & Polarised word restored \\
    \cyr{РоЕ} (Cyr.) & PoE (Lat.) & PoE (Lat.) & \texttt{U+0420 U+043E U+0415} $\rightarrow$ \texttt{U+0050 U+006F U+0045} & Mixed token normalised \\
    \cyr{КМК} 2.01.03-19 & unchanged & \cyr{КМК} 2.01.03-19 & -- & Domain code preserved \\
    LMS.TUIT.UZ & unchanged & LMS.TUIT.UZ & -- & Abbreviation preserved \\
    \bottomrule
  \end{tabular}
  \par\smallskip\footnotesize The first three columns are visually almost identical, which is the nature of the homoglyph problem; the difference appears only at the level of code points. In rows 1--3 the model output matches the reference. In rows 4--5 the raw input is already correct and the Levenshtein bound $\tau$ of Equation~(\ref{eq:conservative}) blocks the rewrite, which is the mechanism that holds newly introduced errors at 84.
\end{table*}

The last two rows carry as much weight as the first three. Both comparison systems fail on exactly such strings: the rule-based corrector does not treat a substituted token as out-of-vocabulary and leaves it untouched, while the LLM reads codes such as \cyr{КМК} 2.01.03-19 or LMS.TUIT.UZ as unknown words and rewrites them, which accounts for a substantial share of its 1,132 new errors.

\subsection{Breakdown by script type and scan quality}
The working assumption was that the advantage of the model would appear mainly where homoglyph confusion is dense. To test it, Table~\ref{tab:breakdown} divides the Gold set into groups by script type and scan quality.

\begin{table}[t]
  \caption{Model performance by text type and scan quality}
  \label{tab:breakdown}
  \begin{tabular}{lrrrr}
    \toprule
    Condition & $N$ & Raw CER & Model CER & ERP \\
    \midrule
    Cyrillic, clean & 1,008 & 0.0608 & 0.0403 & +33.7\% \\
    Cyrillic, noisy & 1,000 & 0.0569 & 0.0365 & +35.9\% \\
    Latin, clean & 1,216 & 0.0610 & 0.0438 & +28.2\% \\
    Latin, noisy & 1,199 & 0.0550 & 0.0374 & +32.0\% \\
    \textbf{Mixed, clean} & \textbf{394} & \textbf{0.0572} & \textbf{0.0311} & \textbf{+45.6\%} \\
    \textbf{Mixed, noisy} & \textbf{387} & \textbf{0.0348} & \textbf{0.0206} & \textbf{+40.8\%} \\
    \bottomrule
  \end{tabular}
\end{table}

On single-script pages the gain ranges from +28.2\% to +35.9\%, stable across alphabet and scan quality. On mixed pages ERP reaches +45.6\% for clean scans and +40.8\% for noisy ones. The \textquotedblleft{}mixed, noisy\textquotedblright{} group produced an unexpected result at first sight: its raw CER is the lowest at 0.0348, although its homoglyph density is almost five times higher than in the other groups. The two figures, however, measure different things. CER covers all types of error, including segmentation errors, lost diacritics and image noise, whereas homoglyph density counts only the errors that arise from alphabet mixing.

Mixed pages come mostly from modern reports prepared digitally and scanned at high quality, so image-related errors are rare while alphabet substitution is frequent.

\subsection{Ablation: the effect of the noise distribution}
To verify that the result depends on the noise matrix rather than on model size, the architecture, the corpus and all hyper-parameters were kept unchanged, and only the noise distribution used in training was replaced (Table~\ref{tab:ablation}).

Random noise that does not match the characteristics of the actual equipment proved not merely useless but harmful: $-75.1\%$ on the synthetic test and $-7.1\%$ on the Gold set. Homoglyph-oriented noise, with the same architecture, gave +24.0\% and +15.1\% respectively. Since only the noise distribution differs between the two runs, it is the principal source of the gap.

\begin{table}[t]
  \caption{Effect of the training noise distribution on CER and ERP}
  \label{tab:ablation}
  \small
  \setlength{\tabcolsep}{3pt}
  \begin{tabular}{lrrrr}
    \toprule
    Training noise & \multicolumn{2}{c}{Synthetic test} & \multicolumn{2}{c}{Gold set} \\
    \cmidrule(lr){2-3}\cmidrule(lr){4-5}
     & CER & ERP & CER & ERP \\
    \midrule
    Random visual noise & 0.1667 & $-75.1\%$ & 0.0619 & $-7.1\%$ \\
    Homoglyph-oriented & 0.0724 & +24.0\% & 0.0491 & +15.1\% \\
    Emp. matrix + domain & n/a & n/a & 0.0369 & +36.2\% \\
    \bottomrule
  \end{tabular}
  \par\smallskip\footnotesize The third configuration reports no synthetic metrics. Its training data were reformatted to follow the report style and rebuilt with a 60\% report / 40\% news ratio, so the original synthetic test is not a comparable benchmark for it.
\end{table}

The third configuration is not a pure ablation: alongside the empirical noise matrix it also uses domain adaptation and conservative decoding, so the increase from +15.1\% to +36.2\% cannot be attributed to the noise distribution alone. Separating the factors would require a full $2 \times 2$ ablation over noise type and domain adaptation, left for future work.

\subsection{Statistical reliability}
Because the Gold set is a single fixed sample, the point estimates carry sampling uncertainty. Confidence intervals were estimated with a non-parametric bootstrap: the line pairs were resampled 1,000 times and the 2.5th and 97.5th percentiles taken as the interval (Table~\ref{tab:bootstrap}). Because the raw and the corrected results are measured on the same lines, the resampling was performed in paired form.

\begin{table}[t]
  \caption{Bootstrap 95\% confidence intervals (5,204 pairs, 1,000 paired resamples)}
  \label{tab:bootstrap}
  \begin{tabular}{lrr}
    \toprule
    Metric & Estimate & 95\% CI \\
    \midrule
    Raw OCR CER & 0.0578 & 0.0565 -- 0.0592 \\
    Model CER & 0.0369 & 0.0358 -- 0.0381 \\
    CER difference (paired) & 0.0209 & 0.0190 -- 0.0218 \\
    ERP & +36.2\% & 34.8\% -- 37.6\% \\
    \bottomrule
  \end{tabular}
\end{table}

\begin{table*}[t]
  \caption{Training configuration, resource usage and deployment}
  \label{tab:config}
  \begin{tabular}{llll}
    \toprule
    Parameter & Value & Parameter & Value \\
    \midrule
    Accelerator & NVIDIA RTX 5090, 32\,GB GDDR7 (Blackwell) & Learning rate & $3 \times 10^{-4}$, linear, 5\% warm-up \\
    Deployment & Single workstation inside the perimeter & Batch / sequence length & $64 \times 256$ bytes \\
    Precision & BF16 mixed precision & Epochs & 16, early stop after 3 \\
    Peak memory & $\approx$26.4\,GB / 32\,GB & Decoding & Beam search, width 4 \\
    Gradient accumulation & Not required & Conservatism threshold & $\tau = 3$ characters \\
    Training time & 12 hours (wall-clock) & Training pairs & 2,283,198 \\
    Data egress & None & Validation / test pairs & 63,420 / 63,426 \\
    Checkpoint & google/byt5-small & Gold evaluation pairs & 5,204 \\
    Optimiser & AdamW, weight decay 0.01 & & \\
    \bottomrule
  \end{tabular}
\end{table*}

The interval for the paired difference does not include zero, so the improvement is not an effect of random sampling, and the ERP interval lies above the M1 threshold of +15\%. The bootstrap estimates variability only within the existing set and does not guarantee generalisation to other collections~\cite{nguyen2019deep}. In addition, some groups in Table~\ref{tab:breakdown}, in particular the \textquotedblleft{}mixed, noisy\textquotedblright{} condition, rest on a small sample of 387 pairs, so the exact ordering has to be interpreted carefully.

\subsection{Computational resources}
Training was carried out on a single workstation inside the perimeter, using an NVIDIA RTX 5090 GPU (32\,GB GDDR7, Blackwell), BF16 mixed precision and the Hugging Face Transformers implementation, for 16 epochs with early stopping; the full configuration is summarised in Table~\ref{tab:config}.

\section{Discussion}
The advantage of the model comes not from general capability but from modelling the region where the errors actually sit. The gap between synthetic and real evaluation in rule-based systems has the same origin: synthetic noise is spread uniformly over characters, whereas real OCR errors concentrate in short tokens with weak context. LLMs capture context well, but sub-word tokenisation prevents them from resolving byte-level corruption, so they rewrite corrupted tokens from general linguistic knowledge. In our first experiments this led to heavy over-correction (CER 0.1219, ERP $-111\%$). The principal limitations of the study follow from the confidentiality requirements: several large baseline models and wider computational resources were not available, which in turn ties the noise matrix to a single document stream and leaves some subsets small. Evaluation is therefore confined to engineering-geological and ICT reports from a single organisation; transfer to legal, literary and administrative texts, whose error profiles are expected to differ, requires separate validation.

\section{Conclusion}
This work addressed post-OCR correction of mixed-script Uzbek text under confidentiality constraints, using a byte-level ByT5 architecture together with an empirical noise generator derived from real errors. With the architecture and the training regime unchanged, replacing the noise distribution alone improved the result sharply, from $-75.1\%$ to +24.0\% on the synthetic test and from $-7.1\%$ to +15.1\% on the Gold set. In the final system the noise distribution, domain adaptation and conservative decoding were applied together; they reduced errors by +36.2\% (95\% CI: 34.8--37.6\%) and met the requirements of the protected environment with 84 new errors. Future work will cover a full ablation separating the noise and domain factors, a comparison of other byte- and character-based architectures under identical noise, and the transfer of the noise matrix to other document streams and Turkic languages.

\end{document}